\documentclass[10pt, journal, compsoc]{IEEEtran}
\IEEEoverridecommandlockouts
\usepackage[english]{babel}
\usepackage[utf8]{inputenc}
\ifCLASSOPTIONcompsoc
\usepackage[nocompress]{cite}
\else
\usepackage{cite}
\fi

\usepackage[hidelinks, pdfauthor={Marco Anisetti, Claudio A. Ardagna, Nicola Bena, Ernesto Damiani},
    pdftitle={Rethinking Certification for Trustworthy Machine Learning-Based Applications}]{hyperref}

\usepackage[english]{babel}
\usepackage[utf8]{inputenc}
\usepackage[T1]{fontenc}
\usepackage[shortcuts]{extdash}
\usepackage{soul, color}
\usepackage{adjustbox}
\usepackage{makecell}
\usepackage{multirow}

\usepackage{pifont}
\usepackage{booktabs}

\usepackage{balance}

\usepackage{tikz}
\usetikzlibrary{arrows.meta, calc, fit, graphs, intersections, positioning, shapes, shapes.misc}

\newcommand{\mathct}{\ensuremath{\mathcal{M}}}

\newcommand{\mathproperty}{\ensuremath{p}}
\newcommand{\mathpropertyall}{\ensuremath{p^*}}

\newcommand{\mathtarget}{\ensuremath{ToC}}
\newcommand{\mathtargetall}{\ensuremath{ToC^*}}

\newcommand{\mathevidcollection}{\ensuremath{\mathcal{E}}}
\newcommand{\mathevidcollectionall}{\ensuremath{\mathcal{E^*}}}

\newcommand{\exmathtarget}[2]{\ifthenelse{\isempty{#2}}{\ensuremath{\mathtarget^{\mathservice{#1}}}}{\ensuremath{\mathtarget^{\mathservice{#1}}}}}
\newcommand{\excm}[1]{\ensuremath{\mathcm_{}}}

\newcommand{\mathservice}{\ensuremath{s}}

\newcommand{\challenge}[1]{C#1}

\newcommand{\mathCertModel}{\ensuremath{\mathcal{CM}}}

\newcommand{\mathCertModelAll}{\ensuremath{\mathcal{CM}^*}}

\newcommand{\factorProcess}{\ensuremath{f_p}}
\newcommand{\factorData}{\ensuremath{f_d}}
\newcommand{\factorModel}{\ensuremath{f_m}}

\newcommand{\ok}{\ding{51}}
\newcommand{\notok}{\ding{55}}

\begin{document}

\title{Rethinking Certification for Trustworthy Machine Learning-Based Applications}

\author{Marco Anisetti, Claudio A. Ardagna, Nicola Bena, Ernesto Damiani
\IEEEcompsocitemizethanks{
    \IEEEcompsocthanksitem Marco Anisetti, Claudio A. Ardagna, Nicola Bena, Ernesto Damiani are with the Department of Computer Science, Università degli Studi di Milano, Milan, Italy. 
\IEEEcompsocthanksitem Ernesto Damiani is alo with with Khalifa University of Science and Technology, Abu Dhabi, UAE. 
    }
}

\markboth{Accepted for publication in IEEE Internet Computing; DOI: \href{https://doi.org/10.1109/MIC.2023.3322327}{10.1109/MIC.2023.3322327}}{DOI: \href{https://doi.org/10.1109/MIC.2023.3322327}{10.1109/MIC.2023.3322327}} 

\IEEEtitleabstractindextext{\begin{abstract}
        Machine Learning (ML) is increasingly used to implement advanced applications with non-deterministic behavior, which operate on the cloud-edge continuum. The pervasive adoption of ML is urgently calling for assurance solutions assessing applications non-functional properties (e.g., fairness, robustness, privacy) with the aim to improve their trustworthiness. Certification has been clearly identified by policymakers, regulators, and industrial stakeholders as the preferred assurance technique to address this pressing need. Unfortunately, existing certification schemes are not immediately applicable to non-deterministic applications built on ML models. This article analyzes the challenges and deficiencies of current certification schemes, discusses open research issues, and proposes a first certification scheme for ML-based applications.
    \end{abstract}
}

\maketitle

Modern applications consist of elastic server\-/side processes running in the cloud, implemented as micro- and nano\-/services developed with cloud-native technologies and orchestrated at run time. The availability of new orchestration platforms and programming frameworks is making it possible to execute these applications at line speed~\cite{montpetit2021computing}.
In the fullness of time, an intelligent cloud continuum will support autonomous, self\-/configuring applications that will request the activation of its services wherever they are needed, including innovative IoT devices~\cite{PRXQuantum.2.017001}. While this new paradigm promises many advantages in terms of availability, elasticity, and, ultimately, quality of service, its complexity is much higher than its predecessors'. This complexity leap is affecting the governance, risk, and compliance landscape, and even the procedures to guarantee security and safety of citizens.

To understand why, let us discuss where we stand right now in terms of available solutions to assess and verify applications behavior\ldots 

Traditionally, assurance techniques have been used to assess and verify the trustworthiness of a system or application~\cite{AADV.CSUR2015}. Along the years, certification has become the most popular assurance technique, providing a way for a trusted authority to assert that a system or application supports a given (set of) non\-/functional property according to some evidence on its operation. Test\-/based certification schemes have been applied to software systems since the Eighties, with the release of the \emph{Orange book}.\footnote{U.S. Department of Defense. ``Department of Defense Trusted Computer System Evaluation Criteria''. Dec. 1985} Test\-/based schemes underwent a crisis in the middle of the past decade, when it became clear that certifying service\-/based applications, even with the low degree of autonomy available at the time, required monitoring and run\-/time re\-/verification, as different services were recruited at each execution. In 2012, on the crest of the service\-/oriented computing wave, the seminal ASSERT4SOA project\footnote{\url{https://cordis.europa.eu/project/id/257351/results}} led the way toward new generation of dynamic certification techniques with the following slogan: \emph{You live in a certified house, you drive a certified car, why would you use an uncertified service?} At that time, certification enabled users to select and compose applications on the basis of their certified properties.

Today, a second crisis of certification schemes is looming, as the massive adoption of machine learning is radically transforming applications behavior~\cite{10.1145/3341145}. 

Opaque ML models pose new concerns on how to test and certify the trustworthiness, safety, and reliability of the applications. The challenges we faced in 2012 are back today, like the most classic of groundhog days. A new slogan is then emerging:

\begin{center}
    \emph{You use certified services, you hire certified professionals, why would you use an application driven by uncertified machine learning?}
\end{center}

This question sheds some light on the motivations of the increasing push towards the definition of sound techniques for the non-functional certification of ML-based applications~\cite{eu-aia,eu-assessment}. What needs to be certified is the \emph{behavior of the application driven by ML models}, rather than some theoretical notion on the ML models.
For instance, fairness means there is no discrimination among users accessing an application; privacy and security mean that the application safely processes user data; robustness means that the application will keep operating reliably while under attack. Nevertheless, these properties depend on the ML models driving the application, and on the process/data used to train them. Training data can be partial or inaccurate (affecting fairness), poisoned (affecting robustness and security), and sensitive (affecting privacy and explainability).

This article outlines the key elements of a sound certification scheme for ML\-/based applications. To this aim, we start from current certification schemes and analyze their limitations. We then give a first definition of a certification scheme for ML\-/based applications, based on simultaneous verification of three factors: \emph{i)} the data used for training, \emph{ii)} the training process, and \emph{iii)} the ML model. We finally apply the scheme in a real\-/world example of its application.

 \section{Traditional Certification Schemes}\label{sec:background-cert}

Traditional certification schemes evolved into dynamic schemes suitable for deterministic composite applications, where services are orchestrated at run time according to their non-functional properties~\cite{AAB.TSC2022}. Application properties are inferred from the ones of its components and continuously verified across service changes~\cite{AADP.TWEB2019, 10.1007/978-3-030-61470-6_25}.

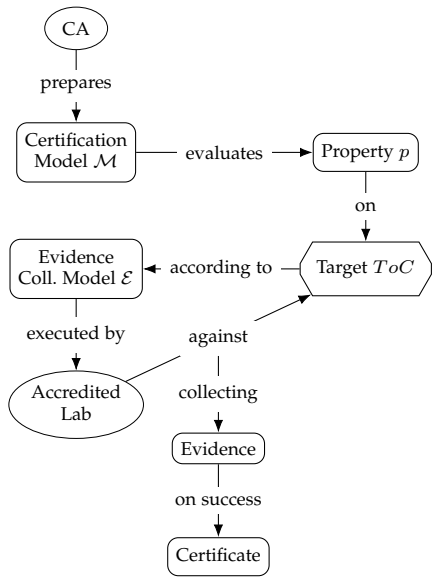
\begin{figure}[t]
  \begin{adjustbox}{max totalsize={\textwidth}{\textheight},center}
      \tikzset{
    entity/.style={
      draw,
      ellipse,
      very thin
    },
    artifact/.style={
      rectangle,
      draw,
      rounded corners,
},
    service/.style={
      chamfered rectangle,
      chamfered rectangle angle=30,
      draw,
},
    every node/.style={
        font={\scriptsize},
        text centered,
    },
    on link/.style={
      fill=white,
    },   
    connector/.style={
        >=stealth,
    },
    connector assurance/.style={
        >=Stealth,
        dashed
    },
    on slide/.code args={<#1>#2}{\only<#1>{\pgfkeysalso{#2}}},
    highlight base/.style={
      very thick
    },
    highlight/.style={
      highlight base,
      fill=LightGoldenRod,
    },
    highlight profile/.style={
      \highlightprofilecolor
    },
    highlight double/.style={
      double=LightGoldenRod
    },
    highlight text/.style={
      text=Goldenrod,
    },
    highlight link/.style={
color=Goldenrod
    },
    highlight light/.style={
      fill=LightGoldenRod!20
    },
      element/.style={
          inner sep=1.5pt,
          font=\footnotesize
      },
      connector/.style={
          >=Stealth,
      },
      lattice layer label/.style={
          text width=2.9cm,
          font={\footnotesize},
      },
      not chosen color/.style={
          black!50
      },
      lattice layer separator/.style={
          dashed,
      },
      service label/.style={
          fill=white,
          circle,
          draw,
          very thick,
          inner sep=2.5pt,
      },
      lattice element bound/.style={
          fill=black!15,
          double
      },
      lattice element selected/.style={
          fill=Goldenrod!15,
          double
      },
      lattice element not chosen/.style={
          not chosen color,
          densely dotted
      },
      lattice element not chosen link/.style={
          not chosen color,
          densely dotted
      },
      service element not chosen/.style={
          text=black!50,
          draw=black!50,
          dashed,
      },
      header/.style={
      },
      main box/.style={
          font={\small}
      },
      link/.style={
        >=Latex,
      },
      activity/.style={
        fill=white
      },
}

\begin{tikzpicture}

    \node[entity] (ca) at (-1.5, 0) {CA};

    \node[artifact] (cm) at (-1.5, -1.65) {\makecell{Certification\\Model \mathct}};

    \draw[->, link] (ca) to node[on link]{prepares} (cm);

    \node[artifact] (properties) at (2.35, -1.65) (property)  {\makecell{Property \mathproperty}};

    \node[service] (target) at (2.35, -3.2) {\makecell{Target \mathtarget}};

    \draw[->, link] (cm) to node[on link] {evaluates} (property);

\draw[->, link] (property) to node[on link] {on} (target);

    \node[artifact] (evidence collection model) at (cm |- target) {\makecell{Evidence\\Coll. Model \mathevidcollection}};

    \node[entity, inner sep=1pt] (lab) at ([yshift=-40]evidence collection model.south) {\makecell{Accredited\\Lab}};

    \draw[->, link] (target) to node[on link] {according to} (evidence collection model);

    \draw[->, link] (evidence collection model) to node[on link]{executed by} (lab);

    \draw[->, link] (lab) to node[on link] (against) {against} (target);

    \node[artifact] at ([yshift=-35]against.south) (evidence) {Evidence};

    \draw[->, link] (against) to node[on link] {collecting} (evidence);

    \node[artifact] at ([yshift=-35]evidence.south) (certificate) {\makecell{Certificate}};

    \draw[ link, ->] (evidence) to node[on link]{on success} (certificate);

\end{tikzpicture}   \end{adjustbox}
  \caption{Traditional certification process. \label{fig:deterministic}}
\end{figure}

Figure~\ref{fig:deterministic} shows the typical process of a certification scheme. The process starts with a Certification Authority (CA) defining the certification model, that is, a specification of the activities to be executed to certify an application. 
Formally, the certification model is a tuple \mathCertModel$=$ $\langle$\mathproperty$,$ \mathtarget$,$ \mathevidcollection$\rangle$, where \mathproperty\ is the \emph{non\-/functional property} to be certified (e.g., confidentiality), \mathtarget\ the \emph{target} of certification, that is, the application to be certified (e.g., a cloud\-/based application), and \mathevidcollection\ the \emph{evidence collection model} (e.g., a set of test cases). An accredited lab, delegated by the CA, executes \mathevidcollection\ to collect the \emph{evidence} that may support the awarding of a \emph{certificate} proving  \mathproperty\ on \mathtarget~\cite{AAB.TSC2022,AADV.CSUR2015}.

Recently, certification schemes broaden the definition of target of certification~\cite{AAB.TSC2022} towards multi-factor certification.

\begin{center}
  \emph{Multi-factor certification is a natural evolution of certification schemes to accomodate the peculiarities of a modern applications. The non-functional posture of an application, in fact, depends on its software artifacts as well as on the processes that brought it to operation (e.g., development and deployment processes).}
\end{center}

However, even dynamic, multi\-/factor certification schemes struggle with the latest evolution of modern applications towards ML. \section{Challenges in ML-Based Applications Certification}\label{sec:challenges}

Today, the certification of ML\-/based applications (ML certification in the following) is more an art than a science~\cite{DA.SOFSEM2020}, resulting in ad hoc solutions tailored to specific properties (e.g., explainability, fairness, and robustness~\cite{AADP.MEDES2020, 10.1145/3375627.3375812, vidot2021certification}). Research is a standstill: no certification scheme for ML is available, despite the increasing push coming from society~\cite{eu-aia}; at the same time, none of the existing system certification schemes~\cite{AADV.CSUR2015} can be adapted to the certification of ML. We argue that this standstill is due to four unsolved challenges we designate here as \challenge{1}--\challenge{4}.

To exemplify the challenges, we introduce a reference scenario that considers a malware detection application (malware detector in short). It is based on a Deep Learning model trained on real data collected from the field, as well as synthetic data generated according to a GAN~\cite{AABGG.MEDES2023}. Data are performance metrics retrieved from the underlying system.
To operate in an adversarial environment, the malware detector must be certified for property \emph{robustness against inference\-/time attacks}, that is, its ability to correctly operate in the presence of adversaries interfering with the malware detection process by perturbing collected data. 

\vspace{.5em}

\noindent \textbf{\challenge{1}: Target definition.} A target of certification \mathtarget\ is commonly defined as a list of components (i.e., endpoints, services, functions) with clear and unambiguous (i.e., deterministic) behavior. This approach does not suit applications whose components are not deterministic~\cite{vidot2021certification}. Current definitions of \mathtarget\ are inapplicable to ML\-/based applications and must evolve to accomplish the uncertainty introduced by ML models. In our reference scenario, the behavior of the deep learner must be certified in terms of the robustness of the dataset and process used for training, as well as the characteristics of the learned model. 

\vspace{.5em}

\noindent \textbf{\challenge{2}: Property definition.} The literature is rich of well\-/formalized non\-/functional properties (e.g., $k$\-/anonymity for privacy, confidentiality, integrity and availability for security), where property definition is decoupled from property verification. The latter is in fact left to the evidence collection model. 
ML certification instead lacks of commonly accepted and rigorous definitions of ML properties (e.g., explainability, fairness, and robustness), where property verification must be included in the property definition. Property verification in fact substantially characterizes the property itself and defines the means driving evidence collection~\cite{AADP.MEDES2020, 10.1145/3375627.3375812}.
For instance, in our reference scenario, property ML robustness must specify how adversarial samples are crafted for run-time verification of the adversarial attacks.

\vspace{.5em}

\noindent \textbf{\challenge{3}: Certification process.} Certification process relies on evidence collection models executing test cases or monitoring rules to collect the evidence at the basis of a certificate award~\cite{AADV.CSUR2015}. Traditional evidence collection statically assesses application interfaces, which might be insufficient to certify the behavior of an ML-based application.
Evidence collection model for ML certification must consider three factors, namely, (training) data, (training) process, and the ML model itself. In particular, factor data is novel and must consider how a model is learned (i.e., developed), including the specific characteristics of the training set. The latter is completely neglected in traditional systems certification and would compare to the evaluation of the application developer (e.g., her experience and skills).

\vspace{.5em}

\noindent \textbf{\challenge{4}: ML pipelines.} The structure of an ML\-/based application is recursive: each of its components can implement an ML pipeline orchestrating other components implementing an aspect of ML, from data ingestion to data processing. Certification schemes must support such a structure. In our reference scenario, the certification of property robustness should consider the robustness of the ML model against inference-time attacks, as well as the integrity of retrieved performance metrics. In some cases, this can be achieved by customizing existing solutions for certificate composition~\cite{AADP.TWEB2019}. However, this challenge remains an open issue that we leave for our future work.
 \section{ML-Based Applications Certification}

To address challenges \challenge{1}--\challenge{3}, we need to reshape traditional certification schemes according to three main aspects: \emph{i) multi-factor certification} of ML\-/based applications behavior (challenge \challenge{1}), \emph{ii) ML-specific non-functional properties} (e.g., fairness, explainability, robustness) definition (challenge \challenge{2}), \emph{iii) ML-specific evidence collection models} supporting non-functional properties verification at point \emph{ii)} (challenge \challenge{3}). 

\vspace{0.5em}

\noindent \textbf{Multi-factor certification of ML-based applications behavior.}
Multi\-/factor certification schemes~\cite{AAB.TSC2022} are the natural choice for certifying ML\-/based applications.
Three factors $f$ should be considered as follows:
\begin{itemize}
    \item \emph{factor data} (\factorData), including information on the dataset used for model training/validation (e.g., characteristics of the samples);
    \item \emph{factor process} (\factorProcess), including information on the training process (e.g., adoption of boosting, transfer learning);
    \item \emph{factor model} (\factorModel), including information on the behavior of the ML model in operation.
\end{itemize}

Our \emph{three-factor} certification model $\mathCertModel$ is a set of three independent certification models $\{$$\mathCertModel^{\factorData},$ $\mathCertModel^{\factorProcess},$ $\mathCertModel^{\factorModel}$$\}$, where each \mathCertModelAll\ is a tuple of the form $\langle$\mathpropertyall,\mathtargetall, \mathevidcollectionall$\rangle$.
Each certification model implements a certification process as follows.

Certification model \mathCertModel$^{\factorData}$$=$$\langle$\mathproperty$^{\factorData}$$,$ \mathtarget$^{\factorData}$$,$ \mathevidcollection$^{\factorData}$$\rangle$ (\emph{factor data}) evaluates the dataset used for training and its impact on the ML model.
For instance, a poisoned training set negatively impacts on property robustness of malware detector, since it reduces its ability to distinguish between benign and malign samples.
\mathCertModel$^{\factorData}$ includes a property \mathproperty$^{\factorData}$ specific for data, a target \mathtarget$^{\factorData}$ modeling the dataset used for training/validation, and an evidence collection model \mathevidcollection$^{\factorData}$ specifying the procedure for collecting evidence on the \mathtarget, including evidence on dataset balancing and feature extraction.

Certification model \mathCertModel$^{\factorProcess}$$=$$\langle$\mathproperty$^{\factorProcess}$$,$ \mathtarget$^{\factorProcess}$$,$ \mathevidcollection$^{\factorProcess}$$\rangle$ (\emph{factor process}) expresses how the training process is implemented and its impact on the ML model. For instance, a sanitization technique for fixing poisoned samples in the dataset positively impacts on property robustness of malware detector, decreasing the shift in the learnt classification boundaries.
$\mathCertModel^{\factorProcess}$ includes a property \mathproperty$^{\factorProcess}$ specific for the training process, a target \mathtarget$^{\factorProcess}$ modeling the training process, and an evidence collection model \mathevidcollection$^{\factorProcess}$ including the procedure for collecting evidence on the design and execution of the training process, such as the inspection of checkpoints generated during training. 

Certification model \mathCertModel$^{\factorModel}$$=$$\langle$\mathproperty$^{\factorModel}$$,$ \mathtarget$^{\factorModel}$$,$ \mathevidcollection$^{\factorModel}$$\rangle$ (\emph{factor model}) evaluates the ML model in operation. 
It is a crucial target of certification and its verification is strongly intertwined with the property to be verified. For instance, property robustness of the malware detector can be actively tested assessing its ability to spot an adversary trying to inject adversarial samples to alter the ML model predictions. As another example, property privacy can be compromised by attacks that infer the presence of a sample in the training set. This is done by inspecting the ML model predictions~\cite{7958568}. However, if the application returns the predicted label only, the attack fails and privacy is preserved despite the lack of a specific protection.
$\mathCertModel^{\factorModel}$ includes a property \mathproperty$^{\factorModel}$ specific for the ML model, a target \mathtarget$^{\factorModel}$ describing the ML model (e.g., its architecture and parameters),  and an evidence collection model \mathevidcollection$^{\factorModel}$ including the procedure for collecting evidence on the behavior of the ML-based application, such as functions exercizing the ML model.

\vspace{0.5em}

\noindent \textbf{ML-specific non-functional properties.}
ML certification requires the definition of ML specific non-functional properties (e.g., fairness, explainability, robustness, safety) and the redesign of traditional non-functional properties (e.g., confidentiality, integrity, availability -- CIA). In general, these properties are the union of factor-specific properties $\mathproperty^{\factorData}$, $\mathproperty^{\factorProcess}$, and $\mathproperty^{\factorModel}$, and must specify the verification means driving evidence collection (see challenge \challenge{2}). We note that, depending on the property, some factors might not be relevant and verification means might be neglected. 

Let us consider property robustness in our reference scenario. It is the union of property robustness of the training set ($\mathproperty^{\factorData}$), the training process ($\mathproperty^{\factorProcess}$), and the ML model ($\mathproperty^{\factorModel}$). 
For example, property robustness of the training set ($\mathproperty^{\factorData}$) is defined as the absence of targeted poisoning in the training set; its definition includes the function used to detect poisoned points.

Let us then consider property integrity. It traditionally proves the integrity of the application and its artifacts. Property integrity of ML instead proves the integrity of the ML model behavior according to the three factors. 
Property integrity in factor data is the integrity of training data (e.g., verifying that the dataset cannot be altered).
Property integrity in factor process is the integrity of the training process (e.g., a training process that includes adversarial training specification).
Property integrity in factor model is the integrity of the generated ML model (e.g., verifying that the packaged ML model is tamper\-/proof).

\vspace{0.5em}

\begin{table*}[!ht]
    \caption{Summary of the certification scheme applied on our scenario.}\label{tbl:exp-summary}
    \begin{adjustbox}{max totalsize={\textwidth}{\textheight},center}
        \begin{tabular}{l p{.39\textwidth} p{.13\linewidth} p{.4\textwidth} c}
            \toprule
            $f$ & \mathproperty & \mathtarget & \mathevidcollection & Outcome\\
            \midrule
            \\
            \factorData &
            \begin{minipage}[t]{\linewidth}
                Robustness against inference\-/time attacks\\
                (Absence of adversarial poisoning in the training set)
            \end{minipage} &
            Training set & 
            \begin{minipage}{\linewidth}
                $-$ Check the presence of poisoned samples\\
                $-$ \emph{evidence}: poisoned samples$=$$\emptyset$
            \end{minipage}&
\ok
            \\
            \\
            \factorProcess &
            \begin{minipage}{\linewidth}
                Robustness against inference\-/time attacks\\
                (Usage of randomized smoothing and adversarial training)
            \end{minipage}&
            Training process &
            \begin{minipage}{\linewidth}
                $-$ Check the training process and checkpoints\\
                $-$ \emph{evidence}: training process does not include \emph{randomized smoothing} and \emph{adversarial training}
            \end{minipage}&
\notok
            \\
            \\
            \factorModel &
            \begin{minipage}[t]{\linewidth}
                Robustness against inference\-/time attacks\\
                (Ineffectiveness of adversarial attacks)
            \end{minipage}&
                Malware detector (Deep Learning model)&
            \begin{minipage}[t]{\linewidth}
                $-$ Craft and send adversarial samples to the ML model \\ $-$ \emph{evidence}: recall$=$0 \end{minipage}&
\notok\\
            \bottomrule
        \end{tabular}
    \end{adjustbox}
\end{table*}
 
\noindent \textbf{ML-specific evidence collection models.} Evidence collection models are factor\-/specific, and describe how to collect evidence according to the three factors. All evidence and metadata collected during the certification of each factor must be stored in certificates, to ensure reproducibility and trustworthiness.

Let us consider property robustness in our reference scenario. \mathevidcollection$^{\factorProcess}$ and \mathevidcollection$^{\factorModel}$ must be adapted to collect data coming from the training process and the ML model coping with their opaqueness. For instance, \mathevidcollection$^{\factorProcess}$ verifies the robustness of the training process by ensuring the usage of strengthening techniques.
\mathevidcollection$^{\factorModel}$ crafts and sends adversarial samples to verify robustness. 

In addition, let us consider an evidence collection process \mathevidcollection$^{\factorModel}$ for property privacy. This property can be verified in terms of the (in)ability to reverse\-/engineer training data while operating the ML model~\cite{7958568, 7536387}.
 \section{Certification in Action}\label{sec:example}

We demonstrate our ML certification scheme in our reference scenario, which considers a malware detection application (see \cite{AABGG.MEDES2023} for more details on the malware detector). We recall that the property of interest is \emph{robustness against inference\-/time attacks}, that is, the ability of the malware detector to behave correctly in the presence of an adversarial perturbation aimed to hide malware activities. Table~\ref{tbl:exp-summary} summarizes the three factors and their outcome. 

\vspace{.5em}

\noindent \textbf{Factor data.} It defines a certification model \mathCertModel$^{\factorData}$$=$$\langle$\mathproperty$^{\factorData}$$,$ \mathtarget$^{\factorData}$$,$ \mathevidcollection$^{\factorData}$$\rangle$ as follows.

\begin{itemize}
    \item Property robustness \mathproperty$^{\factorData}$ is defined as the \emph{absence of poisoned samples from the training set}. They are samples injected in the training set to alter the classification boundaries learnt by the model, thus masking malware activities at inference time~\cite{NEURIPS2018_22722a34}.
    \item Target \mathtarget$^{\factorData}$ is the training set.
    \item Evidence collection model \mathevidcollection$^{\factorData}$\ adapts the poisoning removal technique in~\cite{10.1007/978-3-030-66415-2_4} to flag possibly poisoned samples.
\end{itemize}

In our scenario, no poisoned samples are retrieved, and the assessment for factor data \factorData\ is positive (\ok).

\vspace{.5em}

\noindent \textbf{Factor process.} It defines a certification model \mathCertModel$^{\factorProcess}$$=$$\langle$\mathproperty$^{\factorProcess}$$, $\mathtarget$^{\factorProcess}$$,$ \mathevidcollection$^{\factorProcess}$$\rangle$ as follows.

\begin{itemize}
    \item Property robustness \mathproperty$^{\factorProcess}$\ is defined as the \emph{usage of two strengthening techniques during the training process}: \emph{randomized smoothing} and \emph{adversarial training}. These techniques reduce the presence of poisoned samples and the impact of adversarial samples, respectively.
    \item Target \mathtarget$^{\factorProcess}$ is the training process. It does not include any strengthening techniques.
    \item Evidence collection model \mathevidcollection$^{\factorProcess}$\ inspects the training code, as well as the checkpoints created during training, to retrieve evidence on the usage of \emph{randomized smoothing} and \emph{adversarial training}.
\end{itemize}

In our scenario, the process inspection fails due to the lack of both techniques
and the assessment for factor process \factorProcess\ is negative (\notok).

\vspace{.5em}

\noindent \textbf{Factor model.} It defines a certification model \mathCertModel$^{\factorModel}$$=$$\langle$\mathproperty$^{\factorModel}$$, $\mathtarget$^{\factorModel}$$,$ \mathevidcollection$^{\factorModel}$$\rangle$ as follows.

\begin{itemize}
    \item Property robustness $\mathproperty^{\factorModel}$ is defined as the effectiveness of adversarial attacks against the ML model. It specifies how samples of adversarial attacks are crafted and sent to the model.
    \item Target \mathtarget$^{\factorModel}$ is the malware detector. 
    \item Evidence collection model $\mathevidcollection^{\factorModel}$ exercises the malware detector, sending adversarial (malware) samples and verifying whether the recall retrieved on such samples is larger than $0.95$.
\end{itemize}

In our scenario, recall on adversarial samples is $0$, meaning that they are all misclassified as benign. The assessment for factor model \factorModel\ is negative (\notok).

\vspace{.5em}

To conclude, the malware detector lacks of the proper robustness to operate in adversarial settings. Although it might appear that \factorModel\ is the only relevant factor and traditional certification is sufficient, the certification of the three factors allows the final users to completely understand how the data, the training process, and the model in operation jointly contribute or not to the requested non\-/functional property. For instance, the failure in factor process motivates the failure in factor model. A certificate can be awarded for each factor independently, in case of positive  outcome (\ok), or a single certificate can be released by merging the three assessments according to predefined rules.
 \section{Conclusions}\label{sec:future}
This paper makes a first step towards the certification of ML\-/based applications. The road ahead is still long and impervious.
Our evidence collection techniques based on analysis of the training set, training process, and ML model predictions (Table~\ref{tbl:exp-summary}) must be complemented by other approaches such as \emph{abstract interpretation}~\cite{8418593} or inspection of a surrogate white\-/box model~\cite{10.1145/3236009}.
The life cycle of ML models and their certificates must be carefully managed. ML models need in fact to be continuously adapted according to evolving system behavior, novel requirements, and drift in the incoming data, to name but a few.
This requires certification to follow the changes of ML models along their life cycle. However, certification should not be limited to track changes; it should drive ML models evolution, so that ML models can evolve while supporting the desired properties in all factors. 

To conclude on an optimistic note, we are confident that certification will eventually succeed in supporting trust in ML-based application behavior. It is a good omen that ongoing regulatory discussions~\cite{eu-aia, eu-assessment} agree that certification should target all artifacts involved in ML training and operations.
 
\ifCLASSOPTIONcompsoc
\section*{Acknowledgments}
  The work was partially supported by the projects
  \emph{i)} MUSA -- Multilayered Urban Sustainability Action -- project, funded by the European Union - NextGenerationEU, under the National Recovery and Resilience Plan (NRRP) Mission 4 Component 2 Investment Line 1.5: Strengthening of research structures and creation of R\&D ``innovation ecosystems'', set up of ``territorial leaders in R\&D'' (CUP  G43C22001370007, Code ECS00000037);
  \emph{ii)} SERICS (PE00000014) under the NRRP MUR program funded by the EU -- NextGenerationEU.
\else
\section*{Acknowledgment}
  The work was partially supported by the projects
  \emph{i)} MUSA -- Multilayered Urban Sustainability Action -- project, funded by the European Union - NextGenerationEU, under the National Recovery and Resilience Plan (NRRP) Mission 4 Component 2 Investment Line 1.5: Strengthening of research structures and creation of R\&D ``innovation ecosystems'', set up of ``territorial leaders in R\&D'' (CUP  G43C22001370007, Code ECS00000037);
  \emph{ii)} SERICS (PE00000014) under the NRRP MUR program funded by the EU -- NextGenerationEU.
\fi

\bibliographystyle{IEEEtran}
\bibliography{biblio}

\begin{IEEEbiography}[{\includegraphics[width=1in,height=1.25in,clip]{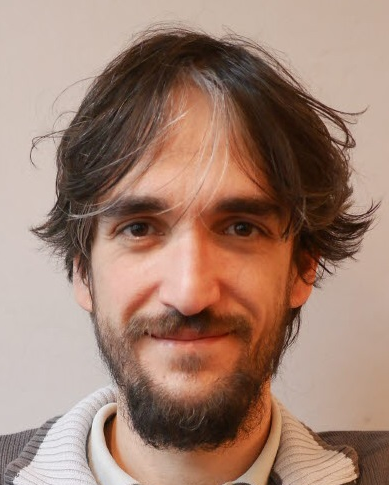}}]{Marco Anisetti}
    is Full Professor at the Università degli Studi di Milano and co-founder of Moon Cloud srl. His research interests are in the area of computational intelligence and its application to the design and evaluation of complex systems. He received the Ph.D. degree in computer science from Università degli Studi di Milano. Contact him at marco.anisetti@unimi.it.
\end{IEEEbiography}

\begin{IEEEbiography}[{\includegraphics[width=1in,height=1.25in,clip]{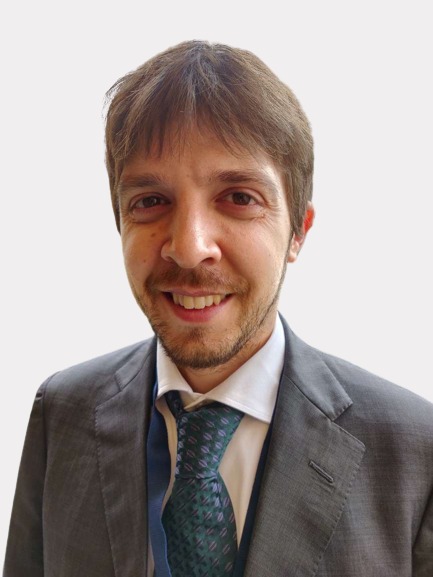}}]{Claudio A. Ardagna}
    is Full Professor at the Università degli Studi di Milano, the Director of the CINI National Lab on Data Science, and co-founder of Moon Cloud srl. His research interests are in the area of cloud-edge security and assurance, and data science. He received the Ph.D. degree in computer science from Università degli Studi di Milano. He is member of the Steering Committee of IEEE TCC, and secretary of the IEEE Technical Committee on Services Computing. Contact him at claudio.ardagna@unimi.it. 
\end{IEEEbiography}

\begin{IEEEbiography}[{\includegraphics[width=1in,height=1.25in,clip]{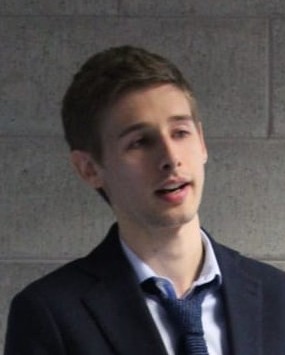}}]{Nicola Bena}
    is a Ph.D. student at the Università degli Studi di Milano. His research interests are in the area of security of modern distributed systems with particular reference to certification, assurance, and risk management techniques. Contact him at nicola.bena@unimi.it. 
\end{IEEEbiography}

\begin{IEEEbiography}[{\includegraphics[width=1in,height=1.25in,clip]{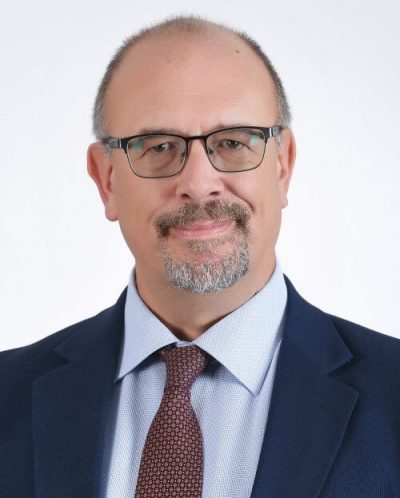}}]{Ernesto Damiani}
    is Full Professor at the Università degli Studi di Milano and Founding Director of the Center for Cyber-Physical Systems, Khalifa University, UAE. His research interests include cybersecurity, big data, and cloud/edge processing. He received an Honorary Doctorate from Institute National des Sciences Appliquées de Lyon, France, in 2017. He is a Distinguished Scientist of ACM and was a recipient of the 2017 Stephen Yau Award. Contact him at ernesto.damiani@ku.ac.ae.
\end{IEEEbiography}

\end{document}